\documentclass{article}

\usepackage[preprint, nonatbib]{neurips_2019}
\newcommand{\G}{\mathcal{G}}

\newcommand{\E}{\mathbb{E}}
\newcommand{\WN}{WN}
\newcommand{\FN}{FN}
\newcommand{\VI}{RAB}

\newcommand{\PS}{\mathcal{V}_{p}}
\newcommand{\ALL}{\mathcal{V}_{\mathrm{all}}}
\newcommand{\RANDOM}{\mathcal{V}_{\mathrm{rand}}}
\newcommand{\ps}{s_p}

\newcommand{\mgenv}{\emph{MultiGoal}}
\newcommand{\sparsemgenv}{\emph{MultiGoal-Sparse}}
\newcommand{\stochmgenv}{\emph{MultiGoal-Stochastic}}
\newcommand{\doorkeyenv}{\emph{Door-Key}}

\usepackage{bmpsize}
\usepackage{bbm}
\usepackage[utf8]{inputenc} % allow utf-8 input
\usepackage[T1]{fontenc}    % use 8-bit T1 fonts
\usepackage{hyperref}       % hyperlinks
\usepackage{url}            % simple URL typesetting
\usepackage{booktabs}       % professional-quality tables
\usepackage{amsfonts}       % blackboard math symbols
\usepackage{nicefrac}       % compact symbols for 1/2, etc.
\usepackage{microtype}      % microtypography
\usepackage{times}
\usepackage{graphicx} % more modern
\usepackage{subfigure} 
\usepackage{caption}
\usepackage{enumerate}
\usepackage{lipsum,multicol}
\usepackage{multirow}
\usepackage[bottom]{footmisc}
\usepackage{enumitem}
\usepackage{hhline}% http://ctan.org/pkg/hhline
\usepackage{xfrac}
\usepackage{array}
\usepackage{bbold}
\usepackage{float}
\usepackage[numbers]{natbib}
\usepackage{amsthm}
\usepackage{amsfonts}
\usepackage{amsmath}
\usepackage{amssymb}
\usepackage{footnote}
\usepackage{braket}
\usepackage{algorithm2e}

\makesavenoteenv{tabular}
\makesavenoteenv{table}
\usepackage{color}
\usepackage[utf8]{inputenc} % allow utf-8 input
\usepackage[T1]{fontenc}    % use 8-bit T1 fonts
\usepackage{hyperref}       % hyperlinks
\usepackage{url}            % simple URL typesetting
\usepackage{booktabs}       % professional-quality tables
\usepackage{amsfonts}       % blackboard math symbols
\usepackage{nicefrac}       % compact symbols for 1/2, etc.
\usepackage{microtype}      % microtypography
\usepackage{makecell}
\usepackage{xcolor}

\title{Learning World Graphs to Accelerate
\\
Hierarchical Reinforcement Learning}

\author{
  Wenling Shang\thanks{: also University of Amsterdam, corresponding at \texttt{w.shang@uva.nl}. $\dagger$: indicates equal contributions. See project page \href{http://www-personal.umich.edu/\textasciitilde shangw/world_graph.html}{(link)} for selected video demos and the Appendix.}\qquad Alex Trott$^\dagger$\qquad  Stephan Zheng$^\dagger$\qquad Caiming Xiong\qquad  Richard Socher\\
  Salesforce Research
}
\begin{document}
\setcitestyle{number}
\maketitle

\begin{abstract}
In many real-world scenarios, an autonomous agent often encounters various tasks within a single complex environment. We propose to build a graph abstraction over the environment structure to accelerate the learning of these tasks. Here, nodes are important points of interest (\emph{pivotal states}) and edges represent feasible traversals between them. Our approach has two stages. First, we jointly train a latent pivotal state model and a curiosity-driven goal-conditioned policy in a task-agnostic manner. Second, provided with the information from the world graph, a high-level Manager quickly finds solution to new tasks and expresses subgoals in reference to pivotal states to a low-level Worker. The Worker can then also leverage the graph to easily traverse to the pivotal states of interest, even across long distance, and explore non-locally. We perform a thorough ablation study to evaluate our approach on a suite of challenging maze tasks, demonstrating significant advantages from the proposed framework over baselines that lack world graph knowledge in terms of performance and efficiency.
\end{abstract}

\vspace{-0.2in}
\section{Introduction}
\label{sec:intro}
\vspace{-0.1in}
%Imagine moving to a new city, everything is unfamiliar at the beginning.
%until we take time to get acquainted with our surroundings. 
%Rather than memorize the exact location of every store or the exact route to navigate between any two points, 
%Once getting acquainted with our surroundings, we build a mental map~\cite{MentalMap} constituting of landmarks, such as a highway junction, a neighborhood's main street, etc, as well as their connecting routes. Thereafter, running other errands become much more manageable. 
%
Many real world scenarios require an autonomous agent to play different roles within a single complex environment.  
For example, Mars rovers carry out scientific objectives ranging from searching for rocks to calibrating orbiting instruments~\cite{Mars}.
%
%Such scenarios give rise to the important research questions of how to form a good understanding of the environment and how to utilize the acquired information for downstream tasks. 
%Under such setups, how to form a good understanding of the environment and then to utilize the acquired knowledge for downstream tasks?
%
Intuitively, a good understanding of the high-level structure of its operational environment would help an agent accomplish its downstream tasks.
%such as the high-level layout, 
%
In reality, however, both acquiring such world knowledge and effectively applying it to solve tasks are often challenging. 
%be able to aid the operation of agents over various tasks, however, neither its formation nor application is trivial in practice.
% 
To address these challenges, we propose a generic two-stage framework that first learns \emph{high-level world structure} in the form of \emph{a simple directed weighted graph}~\cite{graph} and then integrates it into a hierarchical policy model. 

\begin{figure*}[t]
\centering
\includegraphics[width=0.9\textwidth]{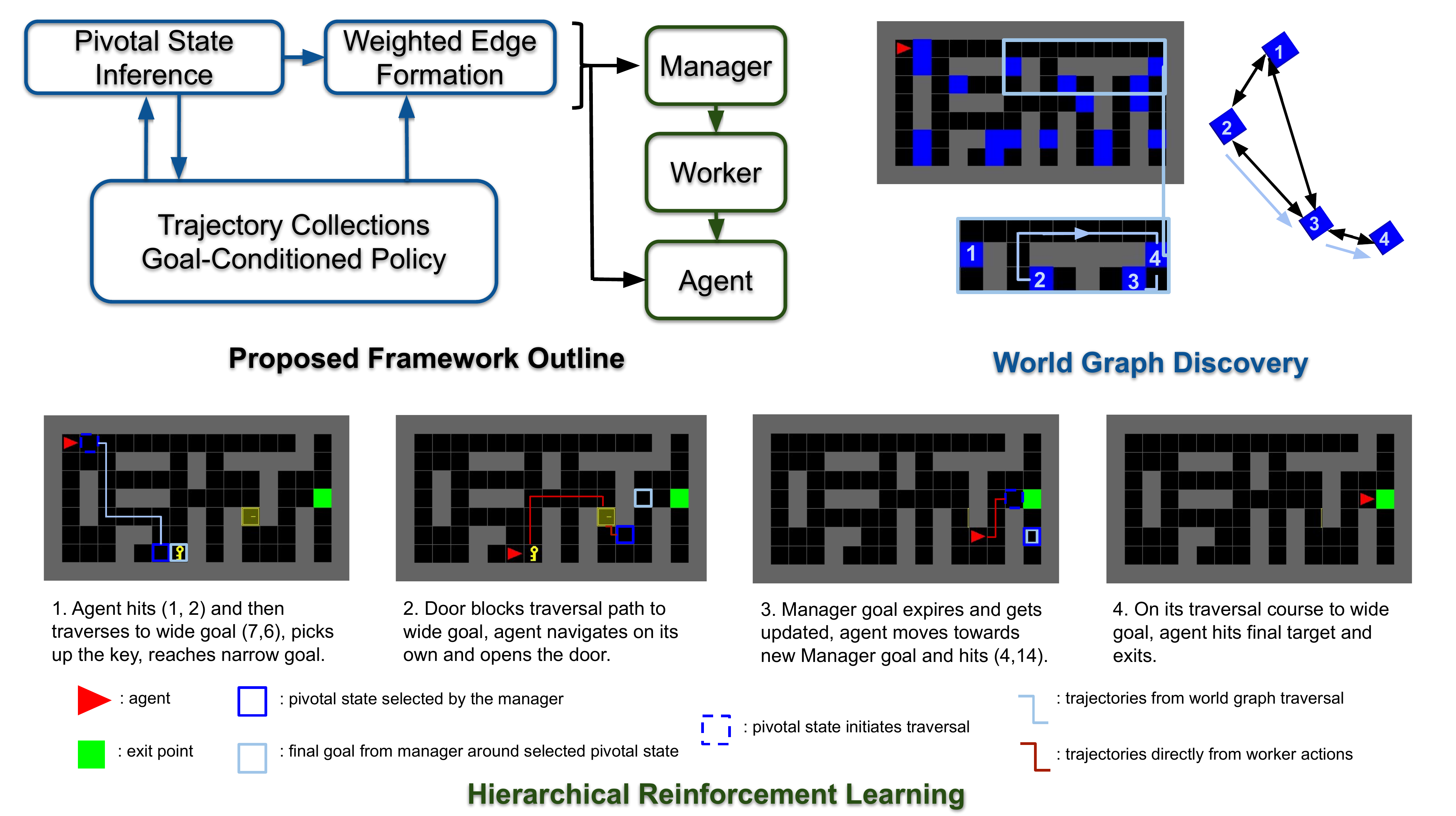}
%\vspace{-0.1in}
\caption{\small{Top Left: Overall pipeline of our proposed 2-stage framework. Top Right (\emph{world graph discovery}): a subgraph exemplifies how to forge edges and traverse between pivotal states (in blue). Bottom (\emph{Hierarhical RL}): an example rollout from our proposed HRL policy with Wide-then-Narrow Manager instructions and world graph traversals, solving a challenging \doorkeyenv{} task.}\label{fig:maze}
}
\vspace{-0.2in}
\label{fig:analysis_action}
\end{figure*}

%In the initial stage, we alternate between devising a progressive and structured exploration strategy for a task agnostic agent to collect meaningful trajectories and optimizing a descriptor of the world, i.e. the environment, in a graph format, referred to as \emph{the world graph}, in an entirely unsupervised fashion. 
In the initial stage, we alternate between exploring the world and updating a descriptor of the world in a graph format~\cite{graph}, referred to as \emph{the world graph} (Figure~\ref{fig:maze}), in an unsupervised fashion. 
%The task-agnostic initial stage aims at establishing an environment graph by guiding an agent to progressively explore and extrapolating information from its trajectories. 
%
%
The nodes, termed \emph{pivotal states}, are the most critical states in recovering action trajectories~\cite{chatzigiorgaki2009real,jayaraman2018time,ghosh2018learning}. 
%
%The nodes termed pivotal states are the most critical states in recovering action trajectories
%
%In particular, given a batch of trajectories, we optimize a fully differentiable recurrent variational auto-encoder~\cite{chung2015recurrent,gregor2015draw,kingma2013auto} with each binary latent variable~\cite{nalisnick2016stick} designated to a specific state, whose learned prior distribution indicates whether it belongs to the set of pivotal states. 
In particular, given a set of trajectories, we optimize a fully differentiable recurrent variational auto-encoder~\cite{chung2015recurrent,gregor2015draw,kingma2013auto} with binary latent variables~\cite{nalisnick2016stick}.
Each binary latent variable is designated to a state and the prior distribution learned conditioning on that state indicates whether it belongs to the set of pivotal states. 
%The nodes correspond to a set of \emph{pivotal states} ({\PS}) within the environment. On a high level, {\PS} is consist of the aforementioned landmarks. Concretely, drawing intuition from video key-frame detection~\cite{chatzigiorgaki2009real,jayaraman2018time} and actionable representation for control~\cite{ghosh2018learning}, we define {\PS} as the most critical states in recovering action trajectories covered by the agent. 
%
%To identify {\PS} from example trajectories, we introduce a novel recurrent auto-encoding model with binary latent variables ({\VI}), whose prior distributions directly indicate how ``pivotal'' their corresponding states are. 
%
%The agent, starting at a currently-identified PS, generates example trajectories from both random walks and a simultaneously learned curiosity-driven goal-conditioned policy~\cite{ghosh2018learning, nair2018visual}, {\goal}. 
%
Wide-ranging and meaningful training trajectories are therefore essential ingredients to the success of the latent model. 
Existing world descriptor learning frameworks often use random~\cite{ha2018world} or curiosity-driven trajectories~\cite{azar2019world}.
%
%Our exploring agent is sent to collect these trajectories via both random walks and a simultaneously learned curiosity-driven goal-conditioned policy~\cite{ghosh2018learning, nair2018visual}. 
Our exploring agent collects trajectories from both random walks and a simultaneously learned curiosity-driven goal-conditioned policy~\cite{ghosh2018learning, nair2018visual}. 
During training, exploration is also initiated from the current set of pivotal states, similar to the ``cells'' in Go-Explore~\cite{ecoffet2019go}, except that ours are learned by the latent model instead of using heuristics. 
%
%Guided by existing knowledge from the world graph, an agent is sent to collect these trajectories via both random walks and a simultaneously learned curiosity-driven goal-conditioned policy~\cite{ghosh2018learning, nair2018visual}. 
%
The edges of the graph, extrapolated from both the trajectories and the goal-conditioned policy, correspond to the actionable transitions between close-by pivotal states. 
Finally, the goal-conditioned policy can be used to further promote transfer learning on downstream tasks~\cite{taylor2009transfer}. 
% assumed to only possess knowledge around its (fixed) starting state initially, is sent to explore the environment in a progressive and structured manner basing on the currently recognized pivotal states as well as a simultaneously learned curiosity-driven goal-conditioned policy~\cite{ghosh2018learning, nair2018visual}. 
%At the very beginning, the agent only possesses knowledge of its initial state and gradually expands its trajectory coverage over the course of alternatively optimizing {\VI}--that is, refining {\PS}--and {\goal}.
%
%Close-by pivotal states are connected by an edge, specifying means to traverse between them, which is extrapolated from the example trajectories. 
%
%The {\PS} discovery is done entirely in an unsupervised fashion, hence this stage is referred to as the {\UPS} module. 
%The diversity of example trajectories and the quality of the PS set improve over the course of alternatively optimizing {\VI} and {\goal}.
%The agent generate example trajectories by both random walk and following a simult
%Our setup assumes the agent at the very beginning only possesses knowledge of its initial state. 
%streaming? refining 

At first glimpse, the world graph seems suitable for model-based RL~\cite{littman1996algorithms,kaiser2019model}, but our method emphasizes the connections among neighboring pivotal states rather than transitions over any arbitrary pair, which is usually considered a much harder problem~\cite{gu2016continuous}.
%
%Explain what it is, then what it relates to
Therefore, in the next stage, we propose a hierarchical reinforcement learning~\cite{kulkarni2016hierarchical,marthi2005concurrent} (HRL) approach to incorporate the world graph for solving specific downstream tasks. 
Concretely, within the paradigm of goal-conditioned HRL~\cite{dayan1993feudal,nachum2018data,vezhnevets2017feudal,levy2017hierarchical}, our approach innovates how the high-level \emph{Manager} provides goals and how the low-level \emph{Worker} navigates.
Instead of sending out a single objective, the Manager first selects a pivotal state from the world graph and then specifies a final goal within a nearby neighborhood of the selected pivotal state. 
We refer to this sequential selection as the \emph{Wide-then-Narrow} ({\WN}) instruction. 
%nspired by recent successes of the goal-conditioned hierarchical reinforcement learning (HRL) paradigm where a high-level manager conveys a goal state for a low-level worker to reach\cite{dayan1993feudal,nachum2018data,vezhnevets2017feudal,levy2017hierarchical}, the next stage proposes a highly effective and general scheme for HRL with the aid of the information extrapolated from the {\UPS} module. 
%
%The manager sequentially outputs 2 goals: first one, chosen from {\PS}, provides a broad direction, which we refer to as the wide goal {$\W$}; the subsequent narrow goal {$\N$} comes from the neighborhood centered around {$\W$} to communicate details. 
%
%In this way, the {\UPS} graph can be nicely put in . 
%
This construction allows us to utilize the information from the learned graph descriptor and to form passages between pivotal states through application of graph traversal techniques~\cite{bertsekas1995dynamic}, thanks to which the Worker can now focus on local objectives.
% In this way, as navigating from its nearby pivotal state to the desired one is greatly simplified thanks to applying graph traversal techniques~\cite{bertsekas1995dynamic} on the world graph, the Worker can focus more on achieving local objectives. the traversal sequences connecting pivotal states through the application of 
%
%
Lastly, as previously mentioned, the goal-conditioned policy derived from learning the world graph can serve as an initialization to the Manager and Worker, allowing fast skill transfer to new tasks as demonstrated by our experiments. 
%In this way, the manager is able to directly navigate the agent between states from {\PS} by traversing along the graph~\cite{bertsekas1995dynamic}, allowing the worker to focus on achieving more local objectives.
%
%Moreover, the worker can be initialized with the goal-conditioned agent obtained from the {\UPS}, as prior works have shown that goal-conditioned policies benefit transfer learning to more difficult tasks~\cite{ghosh2018learning}\wendy{cite another work here}. 
%
%We term the overall second stage Wide-then-Narrow Hierarchical Reinforcement Learning ({\WN}). 
%{\WN} is a general scheme applicable to various HRL algorithms and in this work we demonstrate its efficacy on top of the popular FeUdal Networks~\cite{vezhnevets2017feudal}.

In summary, our main contributions are:
\begin{itemize}[noitemsep,topsep=0pt,leftmargin=0.5cm]
    \item A complete two-stage framework for 1) unsupervised world graph discovery and 2) accelerated HRL by integrating the graph.
    \item The first stage proposes an unsupervised module to learn world graphs, including a novel recurrent differentiable binary latent model and a curiosity-driven goal-conditioned policy. 
    \item The second stage proposes a general HRL scheme with novel components such as the Wide-then-Narrow instruction, navigation via world graph traversal and skill transfer from goal-conditioned policy models.
    \item Quantitative and qualitative empirical findings over a complex 2D maze domain show that our proposed framework 1) produces a graph descriptor representative of the world and 2) improves both sample efficiency and final performance in solving downstream tasks by a large margin over baselines that lack the descriptor.
\end{itemize}
% 
%\begin{itemize}
   % \item Contribution 1
   % \item Contribution 2
    %\item Contribution 3
%\end{itemize}

%shows navigating between these states can be effectively planned by graph traversal such as dynamic programming. 
%We evaluate our approach on 2D maze environments of various sizes and compositions. 
%from the proposed Wide-then-Narrow (WN) HRL with world graph traversal on the Door-Key task. Solid blue bboxes are the pivotal state selected as the ``wide'' goals by the manager and light blue the ``narrow'' goal from local to the pivotal state. Dotted blue bboxes are intermediate pivotal states that initiate traversals to the solid ones, path is colored in light blue; red path is directly executed by the worker. WN-HRL is not only capable of solving challenging tasks but also provides highly interpretable policies.
\vspace{-0.1in}
\section{Environment}
\vspace{-0.1in}
%\stephan{
%Brief section to explain basics of what a ``world'' / MDP means.
%}
%\wendy{Please check the edit}
For ease of clear exposition and scientific control, %we illustrate and assess our framework on 
we choose finite, fully observable yet complex 2D mazes~\cite{gym_minigrid} as our testbeds, i.e. for each state-action pair and their transitions $(s_t,a_t)\to s_{t+1}$, $s_t, s_{t+1}\in \mathcal{S}, a_t\in\mathcal{A}$ are finite. 
% the transition $(s_t,a_t)\to s_{t+1}$ is deterministic, 
%
%
More involved environments can introduce interfering factors, shadowing the effects from the proposed method, e.g. the need of a well-calibrated latent goal space~\cite{higgins2017scan, zach2019hierarchical, nachum2018near}. Section~\ref{sec:conclusion} briefly speculates on extensions of our framework to other environments as future directions.
%Figure~\ref{fig:maze} displays snapshots of the maze layouts used in our work, from small, to medium, to large. 
%
%The small maze is approximately a sub-component of the medium one for the purpose of investigating the impacts from increase in size, while the large one is of very different composition from the other two. 
%
We employ 3 mazes of small, medium and large sizes with varying compositions (see the Appendix for visualization). 
%
%The mazes stay deterministic over each episode, i.e. the transition $(s_t,a_t)\to s_{t+1}$, while we additionally prototype our framework over a stochastic variation to one of the tasks. 
%
Despite being finite and fully observable, these mazes still pose much challenge, especially when the environment becomes large, engages stochasticity,  provides only sparse reward or requires more complicated logic. 
The maze states received by the agent are in the form of bird-eye view matrix representations.
More details on preprocessing are available in the Appendix. 
\begin{figure*}[t]
\centering
\includegraphics[width=0.9\textwidth]{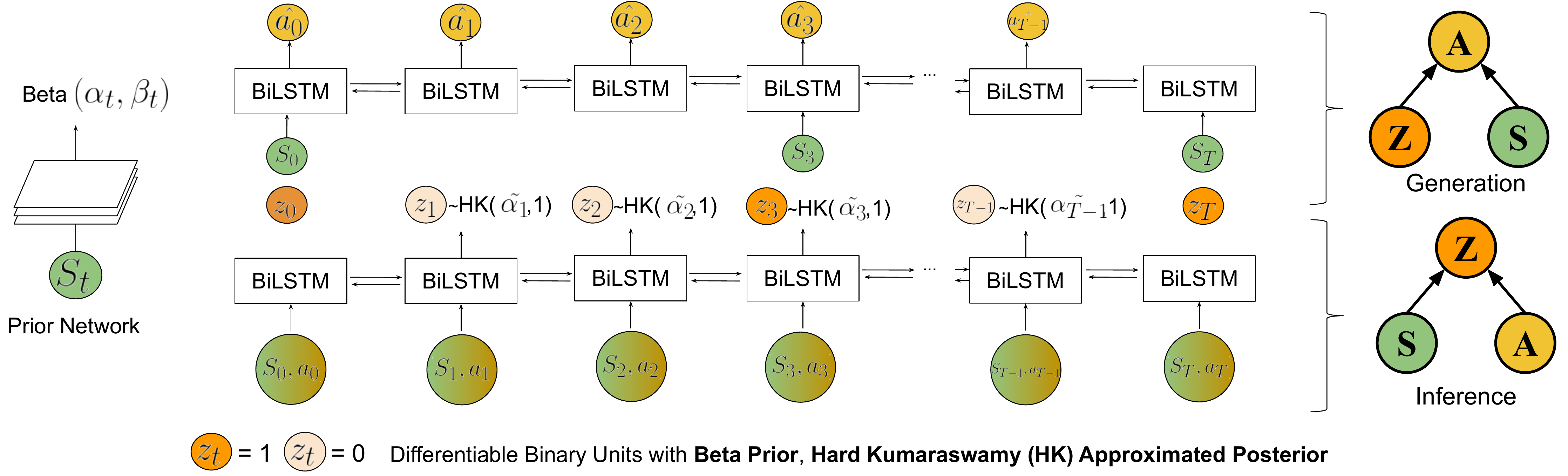}
%\vspace{-0.1in}
\caption{\small{Our recurrent latent model with differentiable binary latent units to discover pivotal states. A prior network (left) learns the state-conditioned prior in Beta distribution, $p_\psi(z_t|s_t) = \mathrm{Beta}(\alpha_t,\beta_t)$. An inference encoder learns an approximate posterior in HardKuma distribution~\cite{basting2019interpretable} inferred from $(s_t,a_t)$'s, $q_\phi(z_t|a_t,s_t) = \mathrm{HardKuma}(\tilde{\alpha_t},1)$. A  generation decoder reconstructs the action sequence from $\{s_t|z_t=1\}$. During training, we sample from $\mathrm{HardKuma}(\tilde{\alpha_t},1)$ using the reparametrization trick~\cite{kingma2013auto}.} \label{fig:GM}
}
\vspace{-0.2in}
\label{fig:analysis_action}
\end{figure*}
%. Learning how to execute
%hortest paths between all pairs of states suggests a deep understanding of the environment dynamics, and we hypothesize that a representation incorporating the knowledge of a goal-conditioned policy can be readily used to accomplish more complex tasks. H
\vspace{-0.1in}
\section{World Graph Discovery}\label{sec:ups}
\vspace{-0.1in}
We envision \emph{a simple directed weighted graph}~\cite{graph} $\G_w$ to capture the high-level structure of the world. Its nodes are a set of points of interest, termed \emph{pivotal states} ($\ps \in \PS$), and edges represent transitions between nodes. 
%
%To construct the graph $\G$ of the world essentially means to identify a set of important landmarks, here we term the pivotal states, as the nodes and how they are connected, i.e. the edges. 
%Pivotal states are defined, inspired by actionable represerntations, as the most necessary states to recover action trajectories. 
Drawing intuition from unsupervised sequence segmentation~\cite{chatzigiorgaki2009real,jayaraman2018time} and imitation learning~\cite{abbeel2004apprenticeship,hussein2017imitation}, we define $\PS$ as the most critical states in recovering action sequences generated by some agent, indicating that these states lead to the most information gain~\cite{azar2019world}.
In other words, given a trajectory $\tau = \{(s_t,a_t)\}_0^T$, we learn to identify the most critical subset of states $\{s_t|s_t \in \PS\}$ for accurately inferring the action sequences taken in the full trajectory $\tau$.
% In other words, given a trajectory $\tau = \{(s_t,a_t)\}_0^T$, only making the subset $\{s_t|s_t \in \PS\}$ from the state sequence visible can well approximate the action sequence taken. 
%actionable representation for control~\cite{ghosh2018learning}, we define it as the most critical states in recovering action trajectories covered by some agent. 

Supposing the state-action trajectories are available, we formulate a recurrent variational inference model (see Section~\ref{sec:VAE})~\cite{blei2017variational,chung2015recurrent,gregor2015draw,kingma2013auto}, treating the action sequences as evidence and, for each state $s_t$ in the sequence, inferring a binary latent variable $z_t$ that controls whether to keep the state for action reconstruction.
%We learn a prior over this inference decision conditioned on only the state (as opposed to the surrounding trajectory) and use its output given state $s$ as the criterion for including $s$ in $\PS$.
We learn a prior over each latent $z_t$ conditioned on its designated state $s_t$, as opposed to using a fixed prior or conditioning on the surrounding trajectory, and use the prior mean as the criterion for including $s_t$ in $\PS$.
%Each state is associated with a learned prior, which at the end measures whether the state belongs to $\PS$. 

Meaningful $\PS$ are learned from meaningful trajectories; hence, we develop a procedure to alternately update the latent model and improve trajectory collections.  
%
%Thus to identify these states,  we send out an agent to explore the environment and to collect trajectories. 
%Then we use a variational recurrent autoencoder with a learned prior for each state from these trajectories to pinpoint which states are in general important to achieve good action sequence reconstruction results. 
When collecting training trajectories, we place the agent at a state from the current iteration's set of $\PS$\textemdash this is possible since the agent can straightforwardly document and reuse the paths from its initial position to states in $\PS$.
This way naturally allows the exploration starting points to expand as the agent discovers more of its environment.
% Imitating real-world setups, a single starting position for the agent is assumed for initial trajectories, from which the latent model learns $\PS$. Now knowing how to reach $\ps$'s, the agent starts rollouts from $\ps\in\PS$, the current points of interest for most effective information gathering. 
% is assumed to only be aware of its 
%The agent initially is only aware of its starting position and hence starts there for each rollout. 
%Then the VI model identifies a bunch of  PSs from these trajectories and along the way the agent record how to reach these pivotal states, and now it can spawn from the updated set of PSs; also because these states are special pivot point, it can be important spots to start from to accelearte exploration. 
%Then alternatively collectiong trajectories and update PS via VI gives us a final set of PS when the reconstruction accuracies no longer decreases. 
%Besides the starting point of exploration, another important ingredient is the policy governing the agent's actions. 
%
Random walk trajectories tend to be noisy thus perhaps irrelevant to real tasks. 
We instead take inspiration from prior work on actionable representations~\cite{ghosh2018learning} and learn a goal-conditioned policy $\pi_g$ for navigating between close-by states, reusing observed trajectories for unsupervised learning (Section~\ref{sec:goal}).
% Inspired by prior works on actionable representations~\cite{ghosh2018learning}, we instead train a goal-conditioned policy $\pi_g$ for navigating between close-by points and reuse its training trajectories as additional source (Section~\ref{sec:goal}).
%
To ensure broad state coverage and diverse trajectories, we add a curiosity reward from the unsupervised action reconstruction error to learn $\pi_g$. 
The latent model is then updated with new trajectories. 
This cycle repeats until the action reconstruction accuracy plateaus. 
To form the edges of $\G_w$ and, we again use both random trajectories and $\pi_g$ (Section~\ref{sec:edge}).
%to To do so, agents are once more spawned from the finalized pivotal states via random walks and $\pi_g: s_p \to s_q \in \PS$ to  forge adjacency between the nodes and their corresponding edge connections (Section~\ref{sec:edge}).
%
Lastly, the implicit knowledge of the world embedded in $\pi_g$ can be further transferred to downstream tasks through weight initialization, which will be discussed later on (Section~\ref{sec:init}).

The pseudo-code summarization of world graph discovery, implementation details, a visualization of how $\PS$ progresses over training and the final $\PS$ from different rollout policies are provided in the Appendix.
% and the model architecture can be found in the code.
The following sections concretely describe each component of our proposed process.
\subsection{Recurrent Variational Model with Differentiable Binary Latent Variables}\label{sec:VAE}
We propose a recurrent variational model with differentiable binary latent variables to discover $\PS$ (Figure~\ref{fig:GM}). 
Given a trajectory $\tau = \{(s_t, a_t)\}_0^T$, we treat the action sequence $\{a_t\}_0^{T-1}$ as evidence in order to infer a sequence of binary latent variables $z_t$'s. %inferred via $(s_t,a_t)$, 
The evidence lower bound is 
 \begin{equation}\label{eq:elbo}
            \mathrm{ELBO} = \mathbb{E}_{q_\phi(\mathbf{Z}|\mathbf{A},\mathbf{S})}\left[\log p_\theta(\mathbf{A}|\mathbf{S},\mathbf{Z})\right] + D_\mathrm{KL}\left(  q_\phi(\mathbf{Z}|\mathbf{A},\mathbf{S})|p_\psi(\mathbf{Z}|\mathbf{S})\right).
    \end{equation}
The objective $\mathbb{E}_{q_\phi(\mathbf{Z}|\mathbf{A},\mathbf{S})}\left[\log p_\theta(\mathbf{A}|\mathbf{S},\mathbf{Z})\right]$ is to reconstruct the action sequence given only the states $s_t$ where $z_t=1$,
% $z_t=1$ indicates that $s_t$ is needed to reconstruct $\{a_t\}_0^{T-1}$.
with the boundary states always given $s_0=s_T=1$. 
To ensure differentiablity, we opt to use a continuous relaxation of discrete binary latent variables by learning a Beta distribution as the priors for $z_t$'s~\cite{russo2018tutorial}.
Moreover, we learn the prior for each $z_t$ conditioned on its associated state $s_t$ (Figure~\ref{fig:GM}).
The prior mean for each $z_t$ signifies \emph{on average} how necessary $s_t$ is for action reconstruction.
% The prior mean for each $z_t$, ($\frac{\alpha_t}{\alpha_t + \beta_t}$), signifies \emph{on average} how necessary to activate $s_i$ for action reconstruction.
%
Also, the KL-divergence term in Equation~\ref{eq:elbo} between the approximated posterior and the learned prior encourages similar trajectories to pick the same states for action reconstruction.
We define $\PS$ as the top 20$\%$ states ranked by the learned prior means.
% We pick the top ($20\%$) $s_t$ based on prior mean ranking to include in $\PS$.  
%The learned Beta prior is conditional on $s_i$ $p(z|s_i)=\mathrm{Beta}(\alpha_i,\beta_i)$ distribution,both indicating \emph{on average} how necessary to keep $s_i$ and encouraging similar trajectories to select the same group of states; at the end of training, we pick the top ($20\%$) states based on the ranking of their prior mean ($\frac{\alpha_i}{\alpha_i + \beta_i}$).
%

The approximate posteriors follow the Hard Kumaraswamy distribution~\cite{basting2019interpretable} [$\mathrm{HardKuma}(\tilde{\alpha}_t, \tilde{\beta}_t)$] 
which resemble the Beta distribution but is outside the exponential family. This choice allows us to sample 0's and 1's without sacrificing differentiability, accomplished via the stretch-and-rectify procedure~\cite{basting2019interpretable,louizos2017learning}. The simple CDF of Kuma also makes the reparameterization trick easily applicble~\cite{kingma2013auto,rezende2014stochastic,maddison2016concrete}.
% % We choose Hard Kumaraswamy distributions\cite{basting2019interpretable}, $\mathrm{HardKuma}(\tilde{\alpha}_t, \tilde{\beta}_t{=}1)$\textemdash a distribution resembling Beta but outside of the exponential family, as approximated posteriors, for it bears the following merits. 
% %
% Although not a concern for prior as it corresponds to the average activation statistics, we do desire $z_t$ to be sampled as 0 or 1 during inference of individual sequences to create an actual binary mask over $\{s_t\}_1^{T-1}$. 
% %
% Thankfully, the stretch-and-rectify procedure~\cite{basting2019interpretable,louizos2017learning} in forming HardKuma from Kuma yields pure 0's and 1's during sampling.
% %
% Secondly, the simplicity of CDF to Kuma distributions is amendable to the reparametrization trick, i.e. refactoring the stochastic unit to a random variable from uniform distribution and another simple differentiable function\cite{kingma2013auto,rezende2014stochastic,maddison2016concrete}.
Lastly, KL-divergence between Kuma and Beta distribution can be approximated in closed form~\cite{nalisnick2016stick}.
We fix $\tilde{\beta}_t=1$ to ease optimization since the Kuma and Beta distributions coincide when $\alpha_i{=}\tilde{\alpha_i}, \beta_i{=} \tilde{\beta}_i{=}1$.
%
%Given a batch of training trajectories, we use a v recurrent model to learn the necessary states to for accurate action trajectory reocnstruction. to sum, we are given a sequence of action states pairs, infer a regularized seuqence of  binary latent variables to include a state or not, then use the included the state to reconstruct the evidence, i.e actions. Mathematically the  GM is in Fig X and variational lower bound is below.
%
%Practically, it consists of 3 parts. Inference, Generation and Prior. 
%Precisely, the inference path takes  a set of trajectories $\tau = \{(s_0, a_0), \cdots (s_T,a_T)\}$ to infer a sequence of binary random variables $\{ z_1, \cdots z_{T-1}\}$\textemdash $z_0=z_T=1$ as we assume the starting and ending states are always known. The latent variables are binary, it indicates whether its corresponding states should be included for reconstruction. Its approximate posterior follow a Kuma distribution for its merit of easy to reparametrize and gives pure zeros and ones. Details see appendix. 

There is not yet any constraint to prevent the model from selecting all states to reconstruct $\{a_t\}_{0}^{T-1}$.
% , a natural but problematic issue as we want to only identify the most representative pivotal states to concisely summarize the world. 
%
To introduce a selection bottleneck, we impose a regularization on the expected $L_0$ norm of $\mathbf{Z}=(z_1\cdots z_{T-1})$ to promote sparsity at a targeted value $\mu_0$~\cite{louizos2017learning, basting2019interpretable}. In other words, this objective constraints that there should be $\mu_0$ of activated $z_t=1$ given a sequence of length $T$. 
Another similarly constructed transition regularization encourages isolated activation of $z_t$, meaning the number of transition between $0$ and $1$ among $z_t$'s should roughly be $2\mu_0$. Note that both expectations in Equation~\ref{eq:reg} have closed forms for HardKuma.
%Borrowing the idea from~\cite{louizos2017learning},
%First is L0 sparsity to ensure not all states are selected. Second is transition regularity to discourage consecutive frames being selected. Lastly, each apprximated posterior is regularized by a prior associated with the state via a KL term. This prior is jointly optimized instead of a fixed one, as we want each state is associated with its own prior, so that the prior can (1) encourage the same state being selected by different trajectoires and (2) the converged prior gives us indication whether this state is pivotal or not. We choose beta prior because its easy to approximate the KL with kuma distribution. 
\begin{equation}\label{eq:reg}
\mathcal{L}_0 = \left\Vert\E_{q_{\phi}(\mathbf{Z}|\mathbf{S},\mathbf{A})}[\|\mathbf{Z}\|_0] - \mu_0\right\Vert^2, \mathcal{L}_{T} = \left\Vert\E_{q_{\phi}(\mathbf{Z}|\mathbf{S},\mathbf{A})}\Sigma_{t=0}^T \mathbbm{1}_{z_t \neq z_{t+1}} - 2\mu_0 \right\Vert^2
\end{equation}
%\mathbf{S},\mathbf{A})}\Sigma_{t=1}^T \mathbbm{1}_{z_t \neq z_{t+1}} - 2\mu_0 \|, \text{ where,}\\
%\E_{q(\mathbf{Z}|\mathbf{S},\mathbf{A})}\left[ \Sigma_{t=1}^T \mathbbm{1}_{z_t \neq z_{t+1}} \right]&= \Sigma_{t=1}^T p(z_t=0)(1-p(z_{t+1}=0) + (1-p(z_{t}=0)(p(z_{t+1} = 0).
%\text{ where }\\
 %\E_{q(\mathbf{Z}|\mathbf{S},\mathbf{A})}[\|\mathbf{Z}\|_0]& =\Sigma_{t=1}^{T} \E_{q(z_t|\mathbf{S},\mathbf{A})}[\mathbbm{1}_{z_t\neq 0}] = \Sigma_{t=1}^{T} 1 - p(z_t = 0|\mathbf{S},\mathbf{A}) = \Sigma_{t=1}^{T} 1 - F_K(\frac{-l}{r-l}; a_t, b_t), \\
\paragraph{ Lagrangian Relaxation.} The overall optimization objective consists of action sequence reconstruction, KL-divergence, $\mathcal{L}_0$ and $\mathcal{L}_T$ (Equation~\ref{eqn:objective}).
% Hyperparameter tuning over the objective weights $\lambda_i$'s is no doubt a daunting chore. 
%
We tune the objective weights $\lambda_i$ using Lagrangian relaxation~\cite{higgins2017beta, basting2019interpretable, bertsekas1999nonlinear}, treating $\lambda_i$'s as learnable parameters and performing alternative optimization between $\lambda_i$'s and the model parameters. 
We observe that as long as their initialization is within a reasonable range, $\lambda_i$'s converge to local optimum autonomously, 
 \begin{equation}\label{eqn:objective}
   \max_{\{\lambda_{1,2,3}\}}\min_{\{\theta,\phi,\psi\}}   \mathbb{E}_{q_\psi(\mathbf{Z}|\mathbf{A},\mathbf{S})}\left[\log p_\theta(\mathbf{A}|\mathbf{S},\mathbf{Z})\right] +\lambda_1 D_\mathrm{KL}\left(  q_\phi(\mathbf{Z}|\mathbf{A},\mathbf{S})|p_\psi(\mathbf{Z}|\mathbf{S})\right) + \lambda_2 \mathcal{L}_0 + \lambda_3 \mathcal{L}_T.
    \end{equation}
Our finalized latent model allows efficient and stable mini-batch training.
 Alternative designs, such as Poisson prior~\cite{kipf2018compositional} for latent space and Transformer~\cite{vaswani2017attention} for sequence modeling, are also possibilities for future investigation.
More details related to the latent model can be found in the Appendix.
%
%More rigorous mathematical model descriptions are in the Appendix and complete implementation details, including hyperparameter and model architecture, can be found in the code. 
%Finally, for generation, we mask out the unselected states, and feed the selected ones to a biLSTM decoder to reconstruct the aciton sequence with softmax. 
\subsection{Curiosity-Driven Goal-Conditioned Agent}\label{sec:goal}
A goal-conditioned policy, $\pi(a_t|s_t,g)$, or $\pi_g$, is trained to reach a goal state $g\in\mathcal{S}$ given current state $s_t$~\cite{ghosh2018learning}.
% ideally leads an agent from its current location to a target goal state $g\in \mathcal{S}$~\cite{ghosh2018learning}. 
% We take standard A2C-LSTM~\cite{wang2016learning,mnih2016asynchronous} (more see Section~\ref{sec:a2c}) to train $\pi_g$ in alternation to latent model optimization.
% 
% 
% 
%, writing the goal $g$ into the input states.
%
%When the environment grows, it can become increasingly difficult to train $\pi_g$ between any two arbitrary states. 
For large state spaces, training a goal-conditioned policy to navigate between any two states is non-trivial.
However, our use-cases, including trajectory generation for unsupervised learning and navigation between nearby pivot states in downstream tasks, only require $\pi_g$ to reach goals over a short range.
% to either generate trajectories or routing between nearby pivotal states (in Section~\ref{sec:edge}).
%n our use case--generating training trajectories and navigating between nearby pivotal states--only asks for $\pi_g$ to excel at traveling relatively locally. 
%
We train such an A2C-based goal-conditioned policy by sampling goals using the end points of random walks with reasonable length from a given starting state.
% To do so, we match each training instance for $\pi_g$ with a random walk trajectory, so that the starting point of the random walk also serves as the current location of the agent and the ending point of the walk is set as the goal state. 
%Thanks to the random walk samples, the same starting states are reused as current locations and ending points as the goal states to maintain tractability.
%
Inspired by the success of intrinsic motivation methods\textemdash in particular, curiosity~\cite{burda2018large,achiam2017surprise,pathak2017curiosity,azar2019world}\textemdash we leverage the readily available action reconstruction errors from the generative decoder as intrinsic reward signals to boost exploration when training $\pi_g$.
The pseudo-code describing this method is in the Appendix. 
\subsection{Edge Connections}\label{sec:edge}
The last crucial step towards the world graph completion is building the edge connections.
After finalizing $\PS$, 
%he edge connections are finalized by deciding the adjacency matrix and path between adjacent $\ps$'s.
we perform random walks from $\ps\in\PS$ to discover the underlying adjacency matrix~\cite{graph} connecting individual $\ps$'s. 
More precisely, we claim a directed edge $s_p\to s_q$ if there exist a random walk trajectory from $s_p$ to $s_q$ that does not intersect a third pivotal state.
We then collect the shortest such actionable paths as the edge paths.
Each path is further refined by $\pi_g$ if feasible.
%, using trajectories collected from the policy when substituting $s_p$ and $s_q$ for the starting state and goal state.
%
The action sequence length of the edge path between adjacent pivotal states defines the \emph{weight} of the edge. 
%For stochastic or partially observable environments, we may entirely count on $\pi_g$ rather than path memorization.
%
Traversal between pivotal states are planned basing on the weight information using dynamic programming~\cite{sutton1998introduction,feng2004dynamic}. 
%Later on when learning downstream tasks, we leverage the weight information to plan traversals among pivotal states. 
%
For deterministic environments, the agents can simply follow the action sequence from the edge to transit between pivotal states. When the environment is stochastic, the agent traverses following the goal-conditioned policy (see Section~\ref{sec:gw}). The planing in this case can potentially be improved by probabilistically or functionally encoding the edge weights~\cite{yamaguchi2016neural,ross2014introduction}, which is left for future work.
\begin{figure*}[t]
\centering
\includegraphics[width=0.99\textwidth]{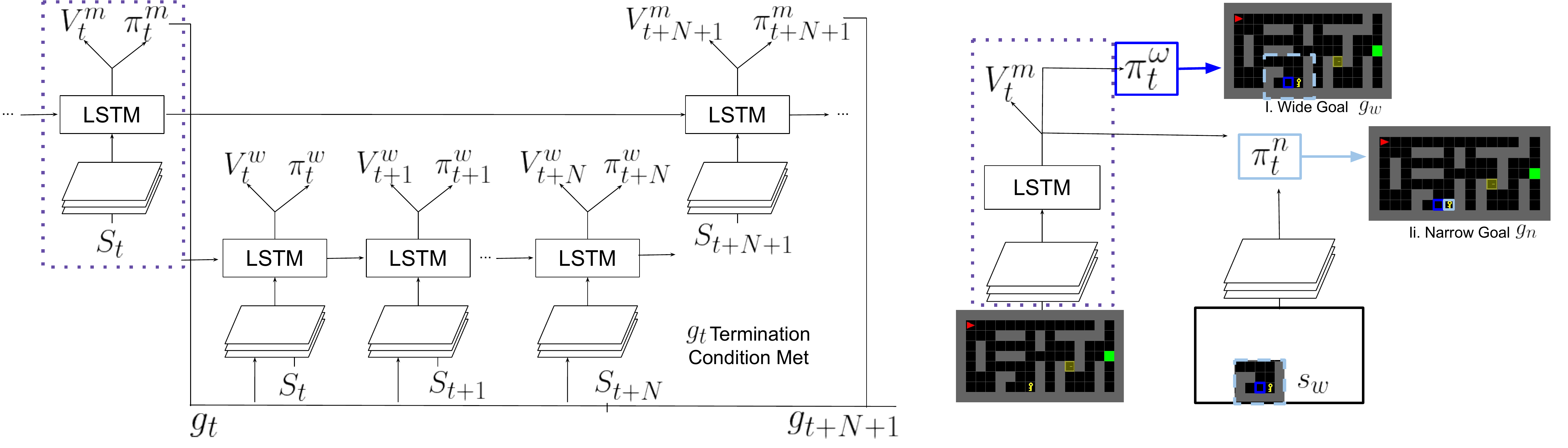}
\vspace{-0.1in}
\caption{\small{Left: a general configuration of Feudal Netowrk; Manager and Worker are both A2C-LSTMs operating at different temporal resolutions. Right: proposed Wide-then-Narrow Manager instruction, where Manager first outputs a wide goal $g_w$ from a pre-defined set of candidate states $\mathcal{V}$, e.g. $\PS$,  and then zooms its attention to a closer up area around $g_w$ to narrow down the final subgoal $g_n$.}\label{fig:task}
}
\vspace{-0.2in}
\end{figure*}
\vspace{-0.1in}
\section{Accelerated Hierarchical Reinforcement Learning}\label{sec:hrl}
\vspace{-0.1in}
We now introduce a hierarchical reinforcement learning~\cite{kulkarni2016hierarchical,marthi2005concurrent} (HRL) framework that leverages the world graph $\G_w$ to accelerate learning downstream tasks.
%The next stage in our framework leverages information attained in Section~\ref{sec:ups} to accelerate Hierarchical Reinforcement Learning (HRL) for specific tasks via several novel components.
%
This framework has three core features: 
\begin{itemize}[noitemsep,topsep=0pt,leftmargin=0.5cm]
\item the Manager uses two-step ``Wide-then-Narrow'' goal descriptions (Section~\ref{sec:wn}),
\item the Worker traverses the learned world graph $\G_w$ at appropriate time(Section~\ref{sec:gw}),
\item the goal-conditioned policy $\pi_g$ learned in the graph discovery stage is used for weight initialization for the Worker and Manager (Section~\ref{sec:init}).
\end{itemize}
%
% Each is generally applicable to many different HRL algorithms.
%These components, namely Wide-then-Narrow manager instruction (Section~\ref{sec:wn}), $\G_w$ traversal (Section~\ref{sec:gw}), and knowledge transfer of $\pi_g$ via initialization (Section~\ref{sec:init}), are applicable to many HRL algorithms.
%
We show that our method learns to solve new tasks significantly faster and better compared to related baselines (Section~\ref{sec:a2c}). 
For implementation details, see the Appendix. 
%This section is gonna talk aboth the 2nd HRL state,esp how we leverage the learning from the previous stage to accelerate this process.  First,we introduce the prelim such as math setup as well as the baseline: Our proposed is essentially a scheme aplicatble to many HRL models, here we choose to demo on top of Fuedual Net, whose individual component are A2Cs. This ection first talks bout the baselines, then the novel component to accelerate the HRL, namely, Wide-to-Narrow, World Traversal and initialization from goal conditioned policy. 
\vspace{-0.1in}
\subsection{Preliminaries and Hierarchical Reinforcement Learning}\label{sec:a2c}
\vspace{-0.1in}
Formally, we consider a Markov Decision Process, where at time $t$ an agent in a state $s_t$ executes an action $a_t$ via a policy $\pi(a_t |s_{t})$ and receives rewards $r_t$. The agent's goal is to maximize its cumulative expected return $R = \E_{(s_t,a_t) \sim \pi, P, P_0}[r_t]$, where $p(s_{t+1} | s_t,a_t), p_0(s_0)$ are the transition and initial state distributions.
%at time $t$ observes $o_t$, which is a function of its state $s_t$, and chooses an action $a_t$ guided by a policy $\pi_t$. Its ultimate objective is to maximize the accumulative expected return over time $R = \E_{(s_t,o_t,a_t)\sim \pi_t}[r_t]$, where $r_t$ is the reward at time $t$. 
%
To solve this problem, we consider a model-free, on-policy learning baseline, the advantage actor-critic (A2C) \citep{wu2016training,pane2016actor,mnih2016asynchronous} and its hierarchical extension called Feudal Network (FN) \cite{dayan1993feudal,vezhnevets2017feudal} (Figure~\ref{fig:task}).
% %%%%
At the core, A2C models both a value function $V(s_t)$ by regressing over the estimated $t_{\max}$-step discounted returns with discount rate $\gamma\in (0,1)$, $\Sigma_{t'=t}^{t_{\max}} \gamma^{t'-t} r_{t'}$, and a policy $\pi$ guided by advantage-based policy gradient~\cite{schulman2015high}. The policy entropy is regularized to encourage exploration.
%, and trains via policy gradients using the experience it collects by sampling from its learning policy, while the value function approximates expected returns. 
%
%The objective for the value network is to estimate the state value $V_t$ by regressing the estimated $t_{\max}$-step discounted returns with discount rate $\gamma\in (0,1)$ (Equation~\ref{eq:value}); the policy network proposes a policy $\pi_t$ and is guided by advantage-based policy gradients using the generalized advantage estimation $\hat{A}$ (details see~\cite{schulman2015high}), regularized by an entropy term to encourage exploration (Equation~\ref{eq:policy}).
% %%%%%
A2C's hierarchical extension FN consists of a high-level controller (``Manager''), which learns to propose subgoals to the low-level controller (``Worker''), which learns to complete the subgoals.
% automatically extrapolate subgoals without manual specification to the low-level controller, i.e. Worker.
%
The Manager receives rewards from the environment and the Worker receives rewards from the Manager by reaching its subgoals.
%based on the actions taken by the Worker 
%
The high and low-level policy models are distinct and operate at different temporal resolutions: the Manager only outputs a new subgoal if either Worker completes its current one or the subgoal horizon $c$ is exceeded.
In this work, we mainly consider finite, discrete and fully observable mazes. As such, FN can use any state as a subgoal and the Manager policy can emit a probability vector of dimension $|\mathcal{S}|$, although our framework supports more general subgoal definitions.
More implementation details and pseudo-codes on our baselines are in the Appendix.
%Actor-Critic based HRL algorithm, in order to better tackle problem such as assignmeing long term credit and sparse reward signamls. FN (with small modifications in our case) has the properties: (1) goal-conditioned, manager gives a goal for worker from the env without manually specify a subpolicy, here our goal is directly a state instead of latent state thanks to fully obervability, as learning latent state add significant difficulty and performance impact independent from the componentw we are proposing. also the manager policy is in the format of softmax--where it output a masked probability heatmask, assigning probability to valid goal states--to allow easy policy gradient formulation. (2)   gradients are
%propagated between Worker and Manager; the Manager receives its learning signal from the environment alone and worker recieves from manager. (3) manager and worker each has its own policy and value but operate over different time scales.  Pipeline see Figure Blah. 
\subsection{World Graph Nodes for Wide-then-Narrow Manager Instructions}\label{sec:wn}
To connect the learned graph $\G_w$ to the HRL framework, the Manager needs the ability to designate any state $s$ as a subgoal while using the abstraction provided by $\G_w$.
To that end, we structure the Manager's output using a  \emph{Wide-then-Narrow} ({\WN}) format:
% to instruct subgoals to Worker in a similar \emph{Wide-then-Narrow} ({\WN}) manner (Figure~\ref{fig:task}). 
%One of our novel components to show signifcant improvement over baseline FN is the introduction of a wide-then-narrow manager instruction style. 
%As shown in Figure~, the manager policy is composed of 2 steps. 

\begin{enumerate}[noitemsep,topsep=0pt,leftmargin=0.5cm]
    \item Given a pre-defined set of candidate states $\mathcal{V}$, the Manager uses a ``wide'' policy $\pi^\omega(s_t)$ that outputs a ``wide'' subgoal $g_w\in\mathcal{V}$. This work proposes the ``wide'' goals come from the learned \textit{pivotal states} $\mathcal{V} = \PS$.
    \item Next, the Manager zooms its attention to an $N\times N$ local area $s_{w,t}{=}s_w(g_{w,t})$ around $g_w$. 
    A ``narrow-goal'' policy $\pi^n$ then selects a final ``narrow'' goal $g_n\in s_{w,t}$, using both global $s_t$ and local $s_{w,t}$ information. 
    Both goals are passed to the Worker, who is rewarded if reaching $g_w$ or $g_n$ within the horizon $c$. 
\end{enumerate}

%
%
%`wide'' subgoal $g_w$ chosen from a set of candidate denoted as $\mathcal{V}$ based on a corresponding wide-goal policy $\pi^w$.
%which serves to quickly direct Worker to the area of interest.
%
%
%coupled with its global vision, emits a ``narrow'' subgoal $g_n$ from the neighborhood according to a conditional policy $\pi^n(\cdot|g_w)$ as the ultimate desired destination for Worker. 
%serving to direct would pick up a state serve as a wide direction, in our case the candidate state set is the set of pivotal states. Then the manager takes a zoomed look of the nbhg of the selected point, taking this into consideration on top of the LSTM info, manager narrow down a more specific goal local to the selected first goal. The agent recieves rewards if it reaches the first goal for the 1st time or the final goal. 
%There are 2 termination conditions, same as baseline Feudal, (1) agent reaches final goal or (2) we set a horizon, and manager exceeds that horizon (example see Figure Blah). 

Using the WN goal format, the policy gradient for the Manager policy $\pi_m$ becomes:
\begin{align*}
&\mathbb{E}_{(s_t,a_t) \sim \pi, p, p_0}\left[ A_{m,t} \nabla \log \left(\pi^\omega\left(g_{w,t}|s_t\right)\pi^n\left(g_{n,t}|s_t, g_{w,t}, s_{w,t} \right)\right) \right] + \nabla \left[\mathcal{H}\left(\pi^\omega\right) + \mathcal{H}\left(\pi^n(\cdot|g_{w,t})\right)\right],
\end{align*}
%  \\
% +\nabla_\theta\mathcal{H}\left(\pi^\omega\right) + \nabla_\theta\mathcal{H}\left(\pi^n(\cdot|g_{w,t})\right).\mathcal{H}(\pi^m) = \sum_{w\in \mathcal{V}}\sum_{w_n \in s_w}\pi^n(w_n|s_w, s_t)\pi^w(w|s_t)\log \pi^n(w_n|s_w, s_t)\pi^w(w|s_t),
where $A_{m,t}$ is the Manager's advantage at time $t$.
% and we regularize the policy entropy to encourage exploration.
% 
Since the size of the action space scales linearly with $|\mathcal{S}|$, the exact entropy for the $\pi^m$ can easily become intractable (see the Appendix). 
% % 
Thus in practice we resort to an effective alternative $\mathcal{H}\left(\pi^\omega\right) + \mathcal{H}\left(\pi^n(\cdot|g_{w,t})\right)$.
% 
% \stz{Wendy: can you check this? what is the approx exactly? I don't think you meant the thing below?}: 
% 
%
% \wendy{the exact one is below, presented in appendix, so we can comment it out here to avoid unnessary confusion?}
% 
% ok fixed
% 
% \begin{align}
% \mathcal{H}(\pi^m) &=  \nabla_\theta\mathcal{H}\left(\pi^\omega\right) + \nabla_\theta\mathcal{H}\left(\pi^n(\cdot|g_{w,t})\right).
% \end{align}
%
% 
% \vspace{-0.2in}
\subsection{Using World Graph Edges for Traversal}\label{sec:gw}
%Another important innovation we add to the HRL stage is allowing the agent to leverate the world graph to the most degree by traversing between pivotal states according to the policy learned from previous stage forming the edges. 
%When treating $\PS$ as the candidate pool for $g_w$'s, 
% 
% 
%one immediate advantage is the utilization of the edges in $\G_w$ and we hereby set forward an example on how.
%
% We hereby set forward an example on how.
%Concretely, the manager sets a pivotal state goal, i..e a node from the graph, during the course of the worker commanding the agent, if it encounters a pivotal state nearby it (and have not reached the assigned pivotal state yet), then it can take advantage of the world graph and dynamic programming to traverse from this one to the desired one,  then workder takes over to continuously search for the local, i.e. 2nd stag goal. 
%In our experiments, 
% 
With pivotal states serving as wide-goals, we can effectively take advantage of the edges in the world graph $\G_w$ through graph traversals:
\begin{enumerate}[noitemsep,topsep=0pt,leftmargin=0.5cm]
    \item \textbf{When to Traverse:} When a Worker is given a goal pair $(g_w, g_n)$, it can traverse the world via $\G_w$ if it encounters a pivotal state $g_w'$ that has a feasible connection to $g_w$ in $\G_w$.
    \item \textbf{Planning:} We estimate the optimal traversal route from $g'_w$ to $g_w$ based on the $\G_w$ edge weights. Here we use the classic dynamic programming planning methods~\cite{sutton1998introduction,feng2004dynamic}, although other (learned) methods can be applied. 
    \item \textbf{Execution:} once the route is planned, for deterministic environments, the agent simply follows the action sequence from the edge paths. For stochastic environments, we either disallow the agent to follow a route that is newly blocked and expect the Manager to adapt accordingly (e.g. in \doorkeyenv{}) or rely on $\pi_g$ to navigate between pivotal states (e.g. in \stochmgenv{}).
\end{enumerate}
During learning, $\pi_g$ can be simultaneously fine-tuned to adapt task-specific environment stochasticity.  
%\stz{Wendy: can you provide details? negative reward?}\wendy{We don't penalize it additionally besides just the step penalty} 
When traversing under $\pi_g$, if the agent fails to reach the next target pivotal state within a certain time limit, it would simply stop its current pursuit.
% 
%We demonstrate this behavior in our experiments. \stz{How?} \wendy{which part??}
%such that there is a path on $\G_w$ to the wide-goal $g_w$, it can navigate from $\ps$ to $g_w$ along the path as if leveraging a documented repertoire of behaviors.
% Besides following step-by-step instructions from Worker, if an agent has not yet reached its current $g_w$ and encounters a pivotal state $\ps\in \PS$ such that there is a path on $\G_w$ in between, it can be immediately transported from $\ps$ to $g_w$ along the traversal path, as if hopping onto a high-speed rail. 
%The agent not only follows step-by-step instructions from Worker, but once encountering a pivotal state $\ps\in\PS$ it is immediately transported to the subgoal $g_w\in\PS$ by traversing $\G_w$, provided that it has not yet reached $g_w$ under the current Manager command horizon and that the path $\ps \to g_w$ is not blocked. 
%
%The optimal traversal route can be estimated basing on edge information via e.g., in our case, dynamic programming~\cite{sutton1998introduction,feng2004dynamic}. 
%
%If a new blockage in the environment (i.e. a door) makes the traversal unviable, we do not allow the Worker to hop onto the blocked path and expect the Manager to learn to plan according to this limitation.
% In the case of blockage, Manager is expected to recognize the situation and plan accordingly or one may enable traversal to a nearby alternative pivotal state. 
%
%
%Incorporating the edges of $\G_w$ in agent maneuver 
The benefit of world graph traversal is to allow the Manager to assign more task-relevant goals that can be far away from an agent's position yet easily reachable by leveraging the connectivity knowledge of the world. 
In this way, we can speed up learning by focusing the low-level exploration on the relevant parts of the world only, i.e., around those highly task-relevant pivotal states $g_w$.
%
% Another foreseeable benefit is the enhancement of exploration, as the agent is no longer restricted to lingering around its current position.
%making slow progress towards other areas. 
%This setup has 2 tremendous benefits. The 1st one is the traversal takes off a lot of learning burdern to the system, which HRL is notorious for training, thanks to the knowledge it gathered from the previous stage. This significantly speed up training. The 2nd one is allowing the worker to focus on more local objecties and outsource any long distance traveling to the world graph--given the manager selects the wide goal appropriately. Thirdly, as the agent as a mean to perform longer and reasonable long distance travelling, it intrinsically is capable of more expansive exploring, which later shown is especially beneficial to task where exploration is essential, such as sparse reward, even without the usage of any other fancy techniques such as curiosity driven or curriculum learning.
\subsection{Knowledge Transfer through Goal-Conditioned Policy Initialization}\label{sec:init}
Lastly, we leverage implicit knowledge of the world acquired by $\pi_g$ during world graph discovery to the subsequent HRL training.
%by reusing its state CNN encoder to initialize Worker and Manager. 
Transferring and generalizing skills between RL tasks often leads to performance gains \cite{taylor2009transfer,barreto2017successor} and goal-conditioned policies have been shown to capture the underlying structure of the environment well~\cite{ghosh2018learning}.
%
%Goal-conditioned policies can capture the underlying structure of the environment and actionable representations derived from such policy can be beneficial for other tasks \cite{ghosh2018learning}.
%
Additionally, optimizing a neural network system like HRL~\cite{co2018self} is sensitive to weight initialization~\cite{mishkin2015all,le2015simple}, due to its complexity and lack of clear supervision. 
Therefore, taking inspiration from the prevailing pre-training procedures in computer vision~\cite{russakovsky2015imagenet,donahue2014decaf} and NLP~\cite{devlin2018bert,radford2019language}, we achieve implicit skill transfer and improved optimization by initializing the task-specific Worker and the Manager with the weights from $\pi_g$.
Our empirical results put forward strong evidence that such initialization serves as an essential basis to solve challenging RL tasks later on, analogously to similar practices in the other domains.
\begin{table*}[t]
\centering
\resizebox{\linewidth}{!}{
\begin{tabular}{lccc|ccc|ccc}
\textbf{Task} & \multicolumn{3}{c}{\textbf{MultiGoal Dense Reward}} & \multicolumn{3}{c}{\textbf{MultiGoal Sparse Reward}} & \multicolumn{3}{c}{\textbf{Stochastic MultiGoal}}   \\

Maze size & Small & Medium & Large & Small & Medium & Large & Small & Medium & Large  \\
\hline 
\textbf{Baselines}  &&&&&&&&& \\
A2C    & $2.04$  &Fail &Fail &$\text{-}0.10$ &Fail &Fail & $1.38$    &Fail &Fail\\
{\FN}     &Fail &Fail &Fail &$0.19$   &Fail &Fail &Fail &Fail &Fail\\
{\FN} $+$ $\pi_g$ init      & $2.93$  &Fail &Fail &Fail &Fail &Fail &$1.93$     &Fail &Fail\\

\textbf{Wide-Narrow}  &&&&&&&&&\\

\textbf{$+$ $\pi_g$-init}  &&&&&& &\\
$\ALL$   & $4.73$ &$4.71$ &Fail &$0.32$   &Fail &Fail &$2.59$  & $1.62$  & Fail\\
$\RANDOM$ & $3.67$ &$4.72$ &Fail &$0.36$  &Fail &Fail &$2.82$   &Fail & Fail \\
{$\PS$}   & $\mathbf{5.25}$ &$\mathbf{5.15}$ &Fail &$0.39$    &Fail &Fail &$\mathbf{3.06}$   & $\mathbf{2.99}$&  Fail\\
  
\textbf{$+$ $\G_w$-traversal}  &&&&&&&\\
$\RANDOM$     & $3.85$ &$2.59$&$1.65$&$0.17$& $0.19$&$0.20$   &Fail&$1.67$& Fail\\
{$\PS$}    & $3.92$ &$2.56$&$2.18$&$0.24$  &$0.20$&$0.16$    &Fail&$2.42$  & $1.05$\\
  
\textbf{$+$ $\pi_g$-init} &&&&&&&&& \\
\textbf{$+$ $\G_w$-traversal}  &&&&&&&&& \\
$\RANDOM$ &$4.16$ &$3.29$&$2.30$&$0.25$  &$0.24$&$0.19$  &$2.72$&$2.42$  & $1.93$\\
{$\PS$} &$5.05$ &$3.00$    &$\mathbf{2.72}$&$\mathbf{0.42}$  &$\mathbf{0.25}$&$\mathbf{0.26}$   &$2.92$&$2.62$& $\mathbf{1.79}$\\

\end{tabular} 
}
\end{table*}

%\caption{
%Results of our control experiments over a ball collecting task on mazes of small, medium and large sizes and dense/sparse reward structures. The average rewards over 3 runs $\pm$ std are reported. 
%Results of our control studies over a ball collecting task varying the size of mazes, the reward structure (dense or sparse), the hierarchical structure, Manager subgoal organization, wide goal candidate,  over  setups for 3 different mazes over 3 different tasks to highlight the effect of using different cnadidate set for wide goals, pretraining with goal conditioned policy and world graph traversal 
%}\label{tab:compare}
\begin{table*}[t]
\centering
\smaller 
\begin{tabular}{lccc|cc|cc}
% \Xhline{2\arrayrulewidth}
\textbf{Door-Key} & \multicolumn{3}{c}{\textbf{$+$ $\pi_g$ init}} &\multicolumn{2}{c}{\textbf{$+$ $\G_w$ traversal}} & \multicolumn{2}{c}{ \textbf{ $+$ $\G_w$ init $+$ traversal}}\\

\textbf{Wide-Narrow}& $\ALL$ &  $\RANDOM$ & $\PS$ &  $\RANDOM$ & $\PS$ &  $\RANDOM$ & $\PS$ \\ 
\hline
Small & $94{\pm}5$&$97{\pm}2$ &$\mathbf{99}{\pm}0$ &Fail&$37{\pm}15$&$76{\pm}14$&$92{\pm}2$\\ 
Medium &$25{\pm}15$&$1{\pm}1$&$56{\pm}2$&Fail&Fail&$\mathbf{79}{\pm}11$&$76{\pm}6$\\
Large &Fail&Fail&Fail&Fail&Fail&$\mathbf{27}{\pm}40$&$\mathbf{26}{\pm}19$\\
% \Xhline{2\arrayrulewidth}
\end{tabular}
\caption{\small{Top: results over \mgenv{} and its sparse/stochastic versions (average reward after 100k training iterations). 
Bottom: results over \doorkeyenv{} (average success rate in $\%$ $\pm$ std), omitting suboptimal models for clearer presentation. 
Note that WN with neither $\pi_g$ init nor $\G_w$ traversal fail across tasks thus not displayed here. \textbf{For full results, including standard-deviations, see the Appendix.}
%``Fail'' means training is either not initiated or validation rewards are never above 0. We omit reporting some suboptimal models on Door-Key for clearer presentation.
%Note that \textbf{all Wide-Narrow baselines  fail to solve the MultiGoal tasks (not displayed here). For full results, including standard-deviations, see the Appendix.}
}}\label{tab:results}
% \vspace{-0.2in}
\end{table*}
% \vspace{-0.1in}
\vspace{-0.1in}
\section{Experiments}
\vspace{-0.1in}
We validate the effectiveness and assess the impact of each proposed component in a thorough ablation study on a set of 4 challenging maze tasks with different reward structures, levels of stochasticity and logic. Furthermore, we evaluate each task in three different mazes of increasing sizes (small, medium and large). Implementation details, snippets of the tasks and mazes are in the Appendix.  

In all tasks, every action taken by the agent receives a negative reward penalty $-0.01$. The other specifics of each task are:
\begin{itemize}[noitemsep,topsep=0pt,leftmargin=0.5cm]
    \item In \mgenv{}, the agent needs to collect 5 randomly spawned balls and exit from a designated exit point. Reaching each ball or the exit point gives reward $+1$. 
    \item Its sparse version, \sparsemgenv{}, only gives a single reward $r\leq 1$ proportional to the number of balls collected upon exiting.
    \item Its stochastic version, \stochmgenv{}, spawns a lava block at a random location each time step that immediately terminates the episode with a negative reward of $-0.5$ if stepped on.
    %and its location randomly changes $25\%$ of the time at each step. 
    \item \doorkeyenv{} is a much more difficult task that adds new actions (``pick'' and ``open'') and new objects to the environment (additional walls, doors, keys).  
    The agent needs to pick up the key, open the door (reward $+1$) and reach the exit point on the other side (reward $+1$).
\end{itemize}
%
% 
%action taken yields a small negative reward to promote the most efficient route.
%It is a rather difficult task with new actions, pick up and open, and changes in the environment that impact $\G_w$, such as additional obstacle to cutting off traversal paths. 
%There are 3 tasks. One is to collect 5 balls spread randomly in the env and then exist, once the agent step on the ball it is considered collected, so no additional action besides moving forward, left and right. Each ball gives reward 1 and exiting give reward one, each action gets a negtative reward -0.01 to encourage the agent to finish fast. we increase this task's difficulty by making the reward sparse, it only receives one reward at exit in proportion to the number of balls it has collected and the maxium reward it recieves is 1 and it still gets step penalties. Third tasks is significantly more challenging where there are additional actions and task logic: the agent needs to find and pick up a key, use it to open a door, then reach the exit point. It only recieves reward signal if it opens the door or reaches the exit point. So MultiGoal < MultiGoal-Sparse < Door-Key. Also the smaller the maze, the easier the task. 

\paragraph{Control Experiments}
We ablate each proposed components and compare against the non-hierarchical and hierarchical baselines, A2C and {\FN}. The proposed components always augment on top of {\FN}.
\begin{itemize}[noitemsep,topsep=0pt,leftmargin=0.5cm]
    \item First, we test initializing the Manager and Worker with the weights of $\pi_g$. 
    \item Next, we evaluate {\WN} with 3 different sets of $\mathcal{V}$'s for the Manager to pick $g_w$ from: $\ALL$ includes all valid states, $\RANDOM$ are uniformly sampled states, $\PS$ are learned pivotal states. $\PS$ and $\RANDOM$ are of the same size and their edge connections are obtained in the same way (Section~\ref{sec:edge})\footnote{The edge connection of $\ALL$ is a trivial case excluded here as every state is 1 step away from its adjacent states. Also, note neither $\pi_g$ nor guaranteed state access is available to  $\RANDOM$ when forming edge connections, but we grant all pre-requisites for the fairest comparison possible.}.
    \item Finally, we enable $\G_w$ traversal on top of {\WN}. If traversal is done through $\pi_g$, along side with HRL training, we also refine $\pi_g$ (if given for initialization) or learn one from scratch (if not given for initialization). 
%\stz{Wendy: i'm not sure why $\pi_g$ is still relevant in the second phase.}  \wendy{for traversal}
\end{itemize}

%with or without hierachiy, and within the hierachical setup, (4) with or without W-N instruction, and within those experiments, (1) different widegoal candidate set, (detail how random is selected!) (2) use pre-trainied initialization or not (3) use graph traversal or not. 

We inherit most hyperparameters from the training of $\pi_g$ in the world graph discovery stage, as the Manager and the Worker both share similar architecture as $\pi_g$.
The hyperparameters of $\pi_g$ in turn follow those from~\cite{shang2018stochastic}.
%except that we lengthen the rollout steps for each iteration from 25 to 60, the amount of iterations from 36K to 50K. 
% 
For details, see the Appendix.
We follow a rigorous evaluation protocol acknowledging the variability in outcomes in deep reinforcement learning \cite{henderson2018deep}: each experiment is repeated with 3 seeds~\cite{wu2017scalable,ostrovski2017count}, 10 additional validation seeds are used to pick the best model which is then tested on 100 testing seeds. Mean of selected testing results are in Table~\ref{tab:results}. We omit results of those experimental setups that consistently fail, meaning training is either not initiated or validation rewards are never above 0, on all tasks. See the Appendix for the full report, including standard deviations over the 3 seeds.  
%
%, the validated best model then is evaluated on 100 reserved testing seeds, and the mean with standard deviation is reported.
% in Table~\ref{tab:results} and all plots. 
%
%\wendy{sufficient training time for failed experiments} We train all the RL models using minibatches of size 32, i.e. 32 agents spawned, with Adam optimizer. Most hyerparameters, such as learning rate, discount factor, rollout per iteration, etc are decided by trials on A2C MultiGoal task for the small maze, both the worker and manager use hyperparameters directly inhired from A2C--as both of them are essentially A2Cs operating on different time scale. Among the HRL specific hyperparameters, the horizon for manager is set to be a third of the rollout steps, the ration of pivotal state is set to be 0.22 and the local zoom-in area for WN instruction is decided basing on this ratio as well as maze size (5x5 for small and medium, 7x7 for large). Worker and manager share similar model architecture which is used by prior works CITE. The subgoal encoder for manager also uses very similar CNN encoding with fewer channels. All details can be found in the code. For evaluation, each experiments we use 3 seeds, standard for DRL (CITE), we plot and report mean + std in figure X and table X. The optimal model for each run is decided by using 10 validation seeds periodically tested during training and the final results are on 100 different testing seeds. For multigoal task, we use the reward as eval metric; for DoorKey, we report success rate, i.e. agent reaching the final exit. 
\subsection{Empirical Analysis}\label{sec:analysis}
%Results are summarized in Table~\ref{tab:results} and Figure XX to XX for the dense reward MultiGoal.
% and we plot the most important control experiment curves in Fig X over MultiGoal, the only non-sparse reward task. 
\paragraph{Initialization with $\pi_g$} 
Table~\ref{tab:results} and Figure~\ref{fig:task} show initialization with $\pi_g$ is crucial across all tasks, especially for the hierarchical models\textemdash e.g. a randomly initialized A2C outperforms a randomly initialized {\FN} on small-maze \mgenv{}. Models starting from scratch fail on almost all tasks within the maximal number of training iterations, unless coupled with $\G_w$ traversal, which is still inferior to using $\pi_g$-initialization. 
%These results also corroborate the claim from \cite{ghosh2018learning} that goal-conditioned policies are promising for task transfer. 
%Figure X shows a randomly initialized A2C outperforms its randomly initialized hierarchical counterpart {\FN}. 
%First it is very easy to see pretrinaing is very important for HRL algorithms, even for the simplest task, where a init from scratch A2C does train to some degree but a non-carefully init HRL would not.  The only way for HRL to train is to use both the proposed WN and world traversal. 
\paragraph{Wide-then-Narrow} 
%However, the init FN is stlil not as good as even A2C, most likely because a vanilla HRL model is notoriously difficult to tame. 
%However, it is seemingly surprising to see how WN HRL trains even when it selects from all valid states for wide goal. 
Comparing A2C, {\FN} and $\ALL$ suggests {\WN} is a highly effective way to structure Manager subgoals. For example, in small \mgenv{}, $\ALL$ ($4.73{\pm}0.5$) surpasses {\FN} ($2.93{\pm}0.74$) by a large margin. We posit that the Manager tends to select $g_w$ from a certain smaller subset of $\mathcal{V}$, simplifying the learning of transitions between $g_w$'s for the Worker. As a result, the Worker can focus on solving local objectives. The same reasoning conceivably explains why $\G_w$ traversal does not yield performance gains on small and medium \mgenv{}. For instance, $\PS$ on small \mgenv{} scores $5.25{\pm} 0.13$, slightly higher than with traversal ($5.05{\pm}0.13$). However once mazes become large, the Worker struggles to master traversals on its own and thus starts to fail the tasks.
%, e.g., on large \mgenv{}. 
%allowing Worker to easily grasp the transition among them. ven in the absence of $\G_w$ traversal
%
%Other potential contributing factors are the additional attention Manager pays to $s_w$ and the 
%selection thanks to the additional attention Manager pays $s_w$ and  It can be of 2 reasons, one is that the zoomed subgoal encoder provides more detailed-scale attention to manager to refine its precise goal decisions; and another possible reason is that that manager, when allowing to select anystate for wide goal, learns to focus on a few major ones and encourage the worker to learn how to move between those (CHEK EXPERIMENTALLY!!!!). In addition, we do find limiting the widegoal candidates to a smaller subset is better than allowing any state. Lastly, by looking at this WN goals, the policies and manager intentions become even more interpretable.  In any case, WN is better than without! 
\paragraph{World Graph Traversal} 
In the case described above, the addition of world graph traversal plays an essential role, e.g. for large \mgenv{}.
As we conjectured in Section~\ref{sec:gw}, this phenomenon can be explained by the much expanded exploration range and a lift of responsibility off the Worker to learn long distance transitions as a result of using $\G_w$ traversal.
%, both thanks to $\G_w$.
%
Moreover, Figure~\ref{fig:task} confirms another conjecture from Section~\ref{sec:gw}: $\G_w$ traversal speeds up convergence, more evidently with larger mazes.
%
%World graph traversal leads to faster convergence, better exploration and long-horizon planning.
%andles challenging tasks such as sparse reward . 
%
%
%Figure~\ref{fig:task} compares validation curves during training on MultiGoal tasks between models with and without $\G_w$ traversal. Those with traversal manifest steeper curves at the beginning of training, more evidently with larger mazes.
%
%In Table~\ref{tab:results}, when the state space is of manageable size and the reward is dense, models without traversal function reasonably well. 
%
%However, once the environment becomes more expansive, e.g. with the large maze, or the reward structure becomes sparse, traversal immediately exhibit much more favorable performances, solving the tasks that are mission-impossible for models without traversal. 
%
Lastly, in \doorkeyenv{}, the agent needs to plan and execute a particular combination of actions. The huge discrepancy on medium \doorkeyenv{} between using traversal or not, $75{\pm}6$ vs $56{\pm}2$, suggests $\G_w$ traversal indeed improves long-horizon planning. 
%Manager in turn adapts to these changes e.g. leading the agent to area close to the door and enforce Worker to learn the sequence of actions to open it thus unblock the path (Figure~\ref{fig:analysis_action}).

\paragraph{Benefit of Learned Pivotal States} 
%Among $\ALL$, $\RANDOM$, and $\PS$, the latter two generally outperform the first one for Manager to select $g_w$ from.
Comparing $\PS$ to $\RANDOM$ reveals the quality of pivotal states identified by the latent model. 
Overall, $\PS$ either performs better (particularly for non-deterministic environments) or similarly as $\RANDOM$, but with much less variance between different seeds.
If one luckily picks a set of random states suitable for a task, it can deliver great results but the opposite is equally possible. 
In addition, edge formation over $\RANDOM$ still depends on the products from learning  world graph, hence using pivotal states with its coupled $\G_w$ is more favorable over $\RANDOM$.
%\wendy{like larger mazes, esp obvious} From our results and plots, the models with traversal converges significantly faster, likely thanks to it seamlessly utilizes traversal knowledge gained from the previous stage's exploration. It also manifests superior exploration property. First, it can enable training even without pretrained initalization, which is the only case that does it, by imposing a wide-range-covered exploration option, the HRL model can overcome the suboptimal init. Second,  even in the face of very sparse reward and difficult task logic, traversal can learn really fast for MultiGoal sparse, and it can solve the DoorKey problem when none of the other controls can. Esp the DoorKey one, we conjecture that the traversal actually implicitly encourages the agent to focus on soving the most difficult component, as when there is no obstacles, the door or the key, the agent can freely travel around, but if the manager gives a wide goal state requires overcoming these obstacles and the traveral path blocks, then the agent would try very hard to conquer the blockage, while not to worry about the easy parts.  
%\paragraph{Impact of learned Pivatol State.} As we observed it is better to limit the wide goal to a subset. Surprisingly, we found uniformally random set works pretty well too. But it doesn't undermine the value of the world discovery stage, as (1) we in fact uses their exploration and goal-conditioned policy to form the traversal graph among the random states, and (2) the runs with random generally display larger variance. 
\begin{figure*}[t]
\centering
\includegraphics[width=0.99\textwidth]{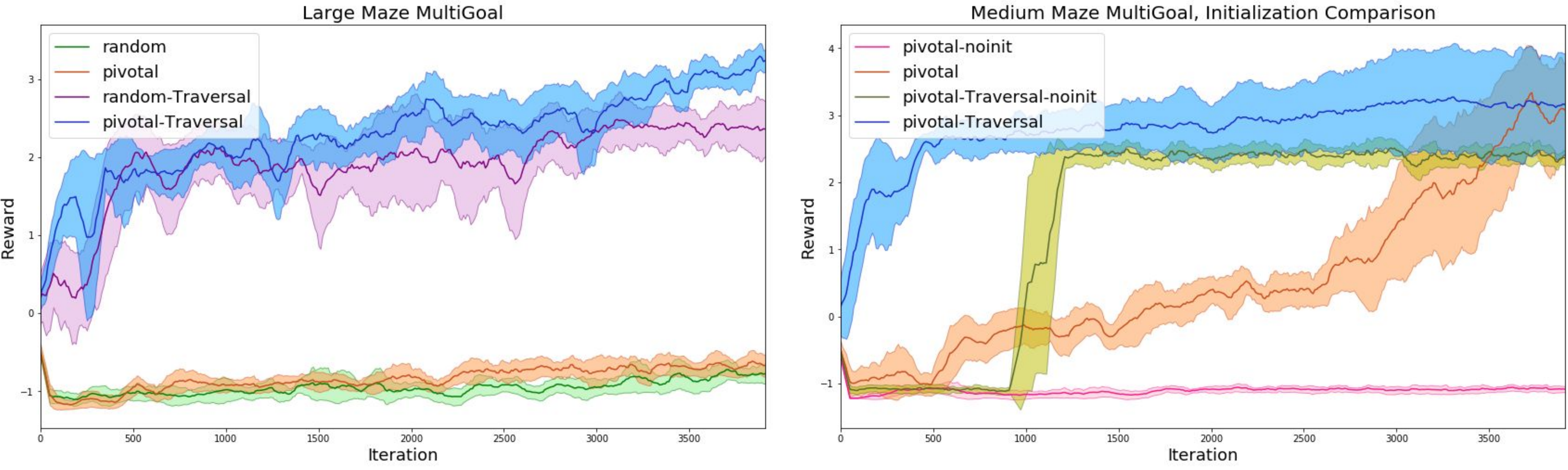}
% \vspace{-0.1in}
\caption{\small{Validation curves during training (mean and standard-deviation of reward, 3 seeds) for \mgenv{}. Left: Compare between $\PS$ and $\RANDOM$, with or without traversal, all models here use {\WN} and $\pi_g$ initialization. Observe that (1) traversal evidently speeds up convergence (2) $\RANDOM$ carries higher variance and slightly inferior performance than $\PS$.  Right: compare with or without $\pi_g$ initialization on $\PS$, all models use {\WN}; initialization shows clear advantage. }\label{fig:task}
}
% \vspace{-0.3in}
\end{figure*}
\vspace{-0.1in}
\section{Related Works}\label{sec:related}
\vspace{-0.1in}
%Our framework covers a variety of topics and thus there are many related works providing inspiration and prior background. 
%
% Our framework intersects with a number of active research areas in generative sequence modeling and hierarchical reinforcement learning. 

Pivotal state discovery is related to unsupervised sequence segmentation~\cite{chatzigiorgaki2009real,jayaraman2018time,pertsch2019keyin,blei2001topic,chan2016latent} and option or sub-task discovery in the context of RL~\cite{jinnai2019discovering, bacon2017option,niekum2013incremental,fox2017multi,leon2016options,kroemer2015towards,kipf2018compositional}. 
Among them, both~\cite{kipf2018compositional} and~\cite{pertsch2019keyin} employ sequential variational models to infer either task boundaries or key frames of demonstrations in an unsupervised manner, followed by applications to hierarchical RL or planning. 
Besides technical differences, instead of turning points for individual sequences, our module aims to identify a set of landmark states that all together can represent the world well. 
Also, our training examples come from carefully orchestrated exploration rather than human demonstrations.
% are not originated from demonstration but well organized exploration.
%The action reconstruction objective is also inspired by imitation learning~\cite{hussein2017imitation,le2018hierarchical,osa2018algorithmic}, i.e. behavior cloning, and actionable representations~\cite{ghosh2018learning}.
%

Understanding the world is central to planning, control and (model-based) RL. 
In robotics, one often needs to locate or navigate itself by interpreting a map of the world~\cite{lowry2015visual,thrun1998learning,angeli2008fast}. 
Our exploration strategy borrows the high-level insight from robotics active localization, where robots are actively guided to investigate unfamiliar regions by humans~\cite{fox1998active,li2016active}. 
Another direction in this area is to learn a world model~\cite{azar2019world,ha2018world,guo2018neural} that generates latent states~\cite{tian2017latent,haarnoja2018latent,racaniere2017imagination}.
%or as parallel auxiliary loss~\cite{oord2018representation,mirowski2016learning}.
%
If the dynamics of the world are also learned, then it can be applied to planning~\cite{mnih2016strategic,hafner2018learning} or model-based RL~\cite{gregor2018temporal,kaiser2019model}.
Although involving generative modeling, our framework differentiates itself through the functionality of our binary latent variables\textemdash indicators of whether a state, regardless of its representation, is a pivotal state.

The policy-learning phase in our framework uses the paradigm of goal-conditioned HRL~\cite{levy2017hierarchical,dayan1993feudal,nachum2018data, vezhnevets2017feudal}. In addition, th {\WN} mechanism borrows ideas from attentive object understanding~\cite{fritz2004attentive,ba2014multiple,you2016image} in vision. 
World graph traversal is inspired by classic optimal planning in Markov Decision Processes with dynamic programming~\cite{bertsekas1995dynamic,feng2004dynamic,weiss1960dynamic,sutton1998introduction}. 
Lastly, initialization with goal-conditioned policies to transfer knowledge is inspired by transfer learning~\cite{donahue2014decaf,taylor2009transfer} and skill generalization in RL~\cite{barreto2017successor,hausman2018learning,ghosh2018learning}.

Concurrently, ~\cite{eysenbach2019search} also proposes to plan a sequence of subgoals leading to a final destination according to a graph abstraction of the world that is obtained via goal-conditioned policy. 
However, under a different problem setup and use-case assumptions,
%\stz{Wendy: what does setup refer to? task?} \wendy{changinge but very similar envirioments for each episode with a single task, they have a single, partially observable env for a single task}, 
the nodes in~\cite{eysenbach2019search} are not learned pivotal states, but directly come from a replay buffer; similarly, their graph is task-specific whereas ours is designed to assist a variety of downstream tasks. 
%Moreover, our work advocates a mastery of navigation between neighboring pivotal states, a much more feasible than than a comprehensive transition model, for downstream RL.

\iffalse
\begin{itemize}
    \item Imitation Learning
    \item actionable representation 
    \item World discovery + curiosity driven intrinsic reward 
    \item Segmentation identification of trajectories; option discovery.
    \item HRL 
    \item Transfer learning: Successor Features for Transfer in Reinforcement Learning 
    \item Go Explore
\end{itemize}
\fi
\vspace{-0.1in}
\section{Conclusion and Future Work}\label{sec:conclusion}
\vspace{-0.1in}
We propose a general two-stage framework to learn a concise world abstraction as a simple directed graph with weighted edges that facilitates HRL for a diversity of downstream tasks. Our thorough ablation studies on several challenging maze tasks show clear advantage of each proposed innovative component in our framework. %

The framework can be extended to other types of environments, such as partially observable and high-dimensional ones, through, e.g., probabilistic or differentiable planning~\cite{kaelbling1998planning,yamaguchi2016neural,ross2014introduction}, latent embedding of (belief) states~\cite{guo2018neural} and goals~\cite{nachum2018near}.
Other directions are to adapt our framework to evolving or constantly changing environments, through, e.g., meta-learning~\cite{finn2017model}, and/or to include off-policy methods to achieve better sample efficiency.
Finally, the learned world graphs can potentially be applied beyond HRL, e.g., multi-tasking RL~\cite{hessel2018multi}, structured exploration by using pivotal state as checkouts~\cite{ecoffet2019go} or in multiagent settings~\cite{bucsoniu2010multi,hu1998multiagent}.
\bibliography{example_paper}
\bibliographystyle{abbrv}
\end{document}